\documentclass[letterpaper]{article} 
\usepackage{aaai2026}
\usepackage{times}  
\usepackage{helvet}  
\usepackage{courier}  
\usepackage[hyphens]{url}  
\usepackage{graphicx} 
\usepackage{natbib}  
\usepackage{caption} 
\usepackage{algorithm}
\usepackage{algorithmic}

\usepackage{newfloat}
\usepackage{listings}
\DeclareCaptionStyle{ruled}{labelfont=normalfont,labelsep=colon,strut=off} 
\floatstyle{ruled}
\newfloat{listing}{tb}{lst}{}
\floatname{listing}{Listing}
\usepackage{amsfonts}
\usepackage{amsmath}
\usepackage{makecell}
\usepackage{enumitem}
\usepackage{tabularx}
\usepackage{array}
\usepackage{booktabs}
\usepackage{caption}

\title{RT-VLM: Re-Thinking Vision Language Model with 4-Clues for Real-World Object Recognition Robustness}

\author {
    Junghyun Park \textsuperscript{\rm 1},
    Tuan Anh Nguyen \textsuperscript{\rm 2},
    Dugki Min \textsuperscript{\rm 1}
}
\affiliations {
    \textsuperscript{\rm 1}Department of Artificial Intelligence, Graduate School, Konkuk University, Seoul, South Korea\\
    \textsuperscript{\rm 2}Department of Software Engineering, Faculty of Information Technology, Ho Chi Minh City University of Industry and Trade (HUIT), Ho Chi Minh City, Vietnam\\
    knlight@konkuk.ac.kr, anhngt@huit.edu.vn, dkmin@konkuk.ac.kr
}

\usepackage{bibentry}

\begin{document}

\maketitle

\begin{abstract}
Real world deployments often expose modern object recognition models to domain shifts that precipitate a severe drop in accuracy. Such shifts encompass (i) variations in low level image statistics, (ii) changes in object pose and viewpoint, (iii) partial occlusion, and (iv) visual confusion across adjacent classes. To mitigate this degradation, we introduce the Re-Thinking Vision Language Model (RT-VLM) framework. The foundation of this framework is a unique synthetic dataset generation pipeline that produces images annotated with "4-Clues": precise bounding boxes, class names, detailed object-level captions, and a comprehensive context-level caption for the entire scene. We then perform parameter efficient supervised tuning of Llama 3.2 11B Vision Instruct on this resource. At inference time, a two stage Re-Thinking scheme is executed: the model first emits its own four clues, then re examines these responses as evidence and iteratively corrects them. Across robustness benchmarks that isolate individual domain shifts, RT-VLM consistently surpasses strong baselines. These findings indicate that the integration of structured multimodal evidence with an explicit self critique loop constitutes a promising route toward reliable and transferable visual understanding.
\end{abstract}

\section{Introduction}

In recent years, deep learning models have attained superhuman accuracy on a range of computer vision benchmarks, and the improvement has been particularly evident in object recognition tasks. Nonetheless, a paradox persists. Models that achieve remarkable scores within controlled training datasets tend to demonstrate fragility when they encounter the unpredictable conditions of deployment. The performance gap between benchmark scenarios and operational robustness stems mainly from "domain shift", the mismatch between the distribution of training data and that of the target environment \citep{shimodaira2000improving}.

This fragility appears in several forms that challenge the dependability of vision systems. For example, a detector regarded as state of the art, despite its high accuracy on common datasets, can be misled by minor variations present in practice \citep{hendrycks2019benchmarking}:

\begin{itemize}
    \item \textbf{Covariate Shift:} Variations in low level image statistics such as blur, noise, or lighting can introduce catastrophic failures after deployment. A model initially trained on clear images can fail to recognize a common object when a moderate blur appears because its learned features are overly specialised to the pristine training distribution \citep{hendrycks2019benchmarking, wang2023survey}. 
    
    \item \textbf{Orientation and Viewpoint Variation:} Objects encountered in natural scenes rarely present themselves in canonical poses familiar to curated datasets. When a network is trained predominantly on frontal views, recognition from an unusual angle can collapse, which indicates a fundamental absence of three dimensional viewpoint invariance in several current architectures \citep{dong2022viewfool}. 
    
    \item \textbf{Occlusion:} In cluttered real world scenes, objects are frequently partially obscured by surrounding clutter. Under incomplete visual evidence the network can fail to detect the target, whereas humans still succeed by exploiting contextual cues \citep{hsiao2012occlusion, wang2020robust}.  
    
    \item \textbf{Class Confusion:} Even with clear and complete views, networks deployed in real settings can confuse visually similar objects. This error demonstrates that the model relies solely on local visual features and disregards the broader scene context that would render the misclassification implausible to a human observer \citep{hendrycks2021natural}.
    
\end{itemize}

Covariate shift, orientation or viewpoint variation, occlusion, and class confusion stand as fundamental barriers to deploying safe and reliable object recognition systems in dynamic scenarios including autonomous driving, robotics, and assistive technologies.

The central hypothesis proposes that visual robustness arises not simply through exposure to larger data collections; rather, it benefits from learning with richer observations coupled with a deliberative reasoning procedure. Instead of emitting a single reflexive prediction, a robust system can gather several sources of evidence and iteratively refine its conclusions in a manner that resembles human cognition.

To explore this hypothesis, the study presents the Re-Thinking Vision Language Model (RT-VLM), a framework that enhances robustness through multimodal self correction. RT-VLM addresses domain shift through two primary innovations:

\begin{enumerate}
    \item \textbf{Learning from Multi Faceted Signals:} A novel synthetic dataset is generated where every image carries the ``4-Clues'' namely bounding boxes, class labels, object level descriptive captions, and a scene level context caption. This dense layered supervision guides the model to ground visual concepts in a rich semantic and contextual web, thereby moving beyond simple object labels.
    \item \textbf{Deliberative ``Re-Thinking'' Inference:} A two stage inference mechanism is proposed that enables the model to perform structured self correction. During the first stage the model produces an initial prediction expressed as the 4-Clues. During the second stage the model treats these clues as internal evidence, responds to an explicit prompt to re think, and corrects earlier mistakes, which improves the final prediction accuracy and reliability.
\end{enumerate}

This paper makes the following primary contributions to the field of robust object recognition:
\begin{enumerate}
    \item A complete and novel pipeline for generating a richly annotated synthetic dataset is introduced. Each image receives four distinct supervisory signals, namely bounding boxes, class names, object level captions, and a context level caption, which are designed to teach models robust and context aware object recognition.
    
    \item A novel two stage inference strategy, termed Re-Thinking, is proposed. This mechanism enables a Vision Language Model (VLM) to perform structured self correction. The model leverages its initial multi faceted output as internal clues to critique and refine subsequent predictions, which elevates accuracy and reduces errors arising from domain shift.
    
    \item Extensive empirical evaluation across challenging benchmarks demonstrates that the combined approach, RT-VLM with Re-Thinking, achieves superior robustness against covariate shift, orientation or viewpoint variation, occlusion, and class confusion, and consistently outperforms baseline systems.
\end{enumerate}

\section{Related Work}

\paragraph{The Limits of Traditional Object Recognition.}
The pursuit of rigorous object recognition has progressed from pure visual pattern matching toward richer multimodal reasoning. Early success in detection can be attributed to Convolutional Neural Networks, whose architectures are grouped into two principal categories: two stage detectors such as RCNN and its successors \citep{ren2015faster}, and one stage detectors exemplified by YOLO \citep{redmon2016you} and RetinaNet \citep{lin2017focal}. The introduction of the Transformer architecture propelled the field further, with DETR \citep{carion2020end} formulating detection as a set prediction problem. Despite solid benchmark numbers, these networks remain vulnerable to domain shift and their accuracy declines when confronted with real world variation in covariates, viewpoint, or occlusion. While domain adaptation methods such as DANN \citep{ganin2016domain} attempt to extract invariant features, they concentrate on the visual modality and frequently leave deeper semantic ambiguities unresolved.

\paragraph{The Multimodal Paradigm Shift: The Rise of Vision Language Models.}
A drive toward stronger robustness is currently provided by Vision Language Models, which learn a shared semantic space between images and text. The influential CLIP model \citep{radford2021learning} illustrated this idea and enabled zero shot classification by aligning images with textual descriptions collected from the web. The concept has expanded rapidly through Multimodal Large Language Models such as LLaVA \citep{liu2023visual}, in which a vision encoder is coupled with a large language model to perform complex visual reasoning. Object detection therefore becomes a language grounding problem, often called Open Vocabulary Detection, where systems including GLIP \citep{li2022glip} and Grounding DINO \citep{liu2023grounding} localise instances described in free form text.

\paragraph{The Frontier of Robustness: Iterative Reasoning and Self Correction.}
As these models grow in capability, recent research has concentrated on equipping them with deliberate stepwise reasoning. Inspiration stems from prompting studies in large language models, for example Chain of Thought \citep{wei2022chain}, which reveal that explicit reasoning improves accuracy. Consequently a line of work on self correction and iterative refinement has emerged. Approaches such as SELF-REFINE \citep{madaan2023self} show that a model can critique and update its own answer using internal feedback. The Re-Thinking mechanism within RT-VLM instantiates this principle for multimodal recognition, overcoming the error detection bottleneck posed by unstructured self correction through a set of four clues generated in the first pass that reconcile location, identity, appearance, and context.

\paragraph{The Practical Enablers: Advanced Datasets and Efficient Fine Tuning.}
The successful deployment of such systems relies on two practical resources, richly annotated data and efficient fine tuning techniques. Dataset development has progressed from simple class labels in PASCAL-VOC \citep{everingham2010pascal} to dense captions in MS-COCO \citep{lin2014microsoft} and Flickr30k Entities \citep{plummer2015flickr30k}. Our four clues dataset, produced with two generative models and one detector, supplies the layered supervision required for advanced reasoning \citep{bauer2024comprehensive}. In parallel, the compute burden of large networks is being tackled by Parameter Efficient Fine Tuning. Low Rank Adaptation \citep{hu2022lora} together with its quantised variant QLoRA \citep{dettmers2023qlora} reduce memory usage by inserting small trainable matrices into frozen backbones, allowing a single graphics processing unit to adapt models with billions of parameters. The RT-VLM framework stands at the intersection of multimodal reasoning, iterative self correction, and pragmatic fine tuning, presenting its main contributions there.

\section{Proposed Method}

\begin{figure*}[t]
\centering
\includegraphics[width=\textwidth]{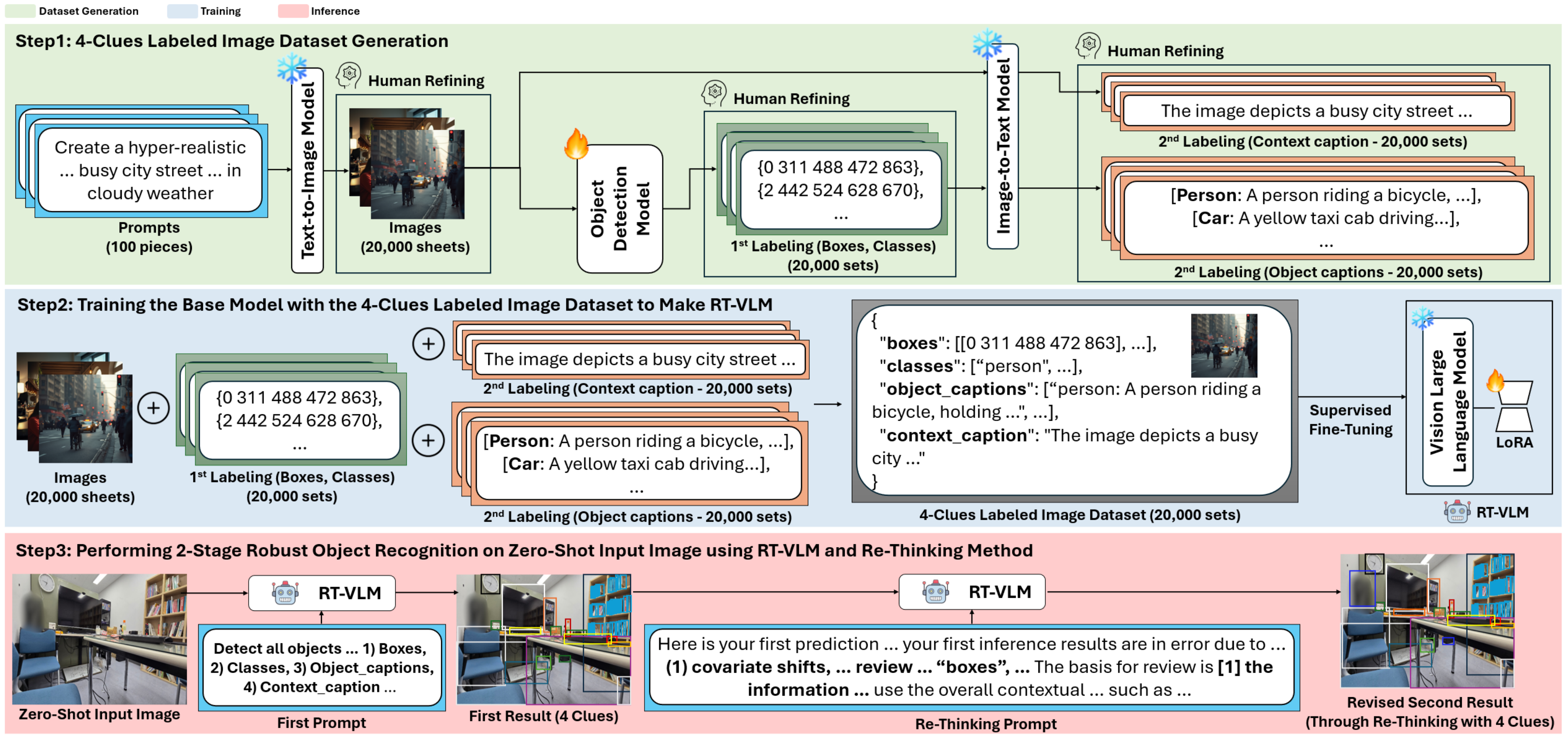}
\caption{Three step pipeline comprising dataset generation, training, and two stage inference.}
\label{fig1}
\end{figure*}

The proposed RT-VLM constitutes a method that spans dataset preparation, model specialisation, and inference refinement. First, a purpose built synthetic corpus is generated. Second, a large scale vision language model is fine tuned on this corpus. Third, inference proceeds in two stages that allow iterative refinement, see Figure \ref{fig1}. The framework runs on a single consumer grade GPU, which indicates a restrained computational footprint.

\subsection{Step1: Four Clues Labeled Image Dataset Generation}

The RT-VLM framework begins with a dataset whose supervisory signal surpasses the information supplied in standard object detection corpora. Our aim is to guide the model toward not only recognising object identity and location, and additionally capturing attributes, state, and contextual relationships. A three model pipeline synthesises images and attaches four distinct clues to every sample.

\subsubsection{Image Generation}

To obtain a wide ranging and demanding corpus, we produce 20{,}000 synthetic images using the FLUX.1‑dev text to image generator \citep{blackforestlabs2025flux}. A structured prompting protocol elicits real world scenarios that exercise the four targeted robustness challenges. Each prompt comprises three building blocks, namely a Pinned Prompt that secures photo realism, one of 100 Base Prompts that describe scenes and objects, and one of 20 Random Modifiers that adjust camera angle, weather, or lighting.

Through this protocol, 100 unique prompt triplets are formed. For every triplet, 200 images are rendered, leading to a corpus of 20{,}000 samples that explicitly encode the visual conditions the downstream model needs to handle. The pre trained FLUX.1‑dev generator runs in 4 bit quantised mode to limit memory demand. A manual verification stage discards defective renders and regenerates replacements to stabilise corpus quality at 20{,}000 images.

\subsubsection{Localization and Classification Labeling}

After image synthesis, we require bounding boxes and class identifiers for every object instance. The task is addressed with a semi automated loop that involves human supervision. We draw a random subset of 1{,}000 images from the corpus, manually mark their bounding boxes and class labels, then split the subset into 700 training, 150 validation, and 150 test images to adapt a pre trained YOLO12x detector \citep{tian2025yolov12}. The adaptation employs a composite objective that combines SIOU regression loss \citep{gevorgyan2022siou}, Distribution Focal Loss \citep{li2020generalized}, and Focal Loss for classification \citep{lin2017focal}. Following 400 training epochs, the model records mAP@0.5 of 0.83, mAP@0.5:0.95 of 0.60, Precision 0.88, Recall 0.75, Intersection over Union mean 0.63, and F1 score 0.81 on the held out test set, which indicates reliable object detection in the synthetic domain. The adapted detector then annotates the remaining 19{,}000 images automatically. A closing manual audit safeguards the correctness of the full set of 20{,}000 labelled images. (Refer to Equations (1) to (4) and Tables 1 and 2.)

\begin{equation}
\mathcal{L}_{SIOU} = C_{IOU}+\frac{\Lambda+\Delta+\Omega}{2}
\label{loss1}
\end{equation}

\begin{equation}
\mathcal{L}_{DFL} = \sum_k FL(q_k,q_k^*)
\label{loss2}
\end{equation}

\begin{equation}
\mathcal{L}_{Focal} = -\alpha(1-p_t)^\gamma logp_t
\label{loss3}
\end{equation}

\begin{equation}
\mathcal{L}_{total} = \lambda_{\text{box}}\mathcal{L}_{SIOU}+\lambda_{\text{dfl}}\mathcal{L}_{DFL}+\lambda_{\text{cls}}\mathcal{L}_{Focal}
\label{loss4}
\end{equation}

with hyperparameters $\alpha=0.25,\ \gamma=1.5,\ \lambda_{\text{box}}=7.5,\ \lambda_{\text{dfl}}=1.5,\ \lambda_{\text{cls}}=0.5$.

\begin{table}[t]
\centering
\renewcommand{\arraystretch}{1.1}
\setlength{\tabcolsep}{5pt}
\begin{tabular*}{\linewidth}{@{\extracolsep{\fill}} c | c | c | c | c @{}}
    \hline
    & SIOU loss & DFL loss & Focal loss & Total loss \\
    \hline
    Initial & 1.21 & 0.27 & 0.19 & 9.58 \\
    \hline
    Final & 0.03 & 0.02 & 0.05 & 0.28 \\
    \hline
\end{tabular*}
\caption{Decrease in training loss of the YOLO12x model.}
\label{table1}
\end{table}

\begin{table}[t]
\centering
\renewcommand{\arraystretch}{1.1}
\setlength{\tabcolsep}{5pt}
\begin{tabular*}{\linewidth}{@{\extracolsep{\fill}} c | c | c | c | c @{}}
    \hline
    & SIOU loss & DFL loss & Focal loss & Total loss \\
    \hline
    Initial & 1.40 & 0.27 & 0.21 & 11.01 \\
    \hline
    Final & 0.05 & 0.04 & 0.05 & 0.46 \\
    \hline
\end{tabular*}
\caption{Decrease in validation loss of the YOLO12x model.}
\label{table2}
\end{table}

\subsubsection{Semantic Labeling: The Final Two Clues.}

Class labels together with bounding boxes convey structural information; however, they convey limited semantics. To enhance meaning, we append two textual layers created by the Janus Pro 7B image to text model \citep{deepseek2025januspro} that is executed with 4 bit quantization.

\begin{enumerate}
\item \textbf{Object Level Captions:} Each detected region is cropped then presented to the caption generator with the prompt ``Describe the object.'' The model returns a short phrase that specifies local attributes, for example ``A person riding a bicycle, holding an umbrella''.
\item \textbf{Context Level Caption:} The uncropped frame is passed to the same generator with the prompt ``Describe the overall scene of this image.'' The resulting sentence outlines the global setting, for instance ``The image depicts a busy city street scene during what appears to be a rainy or overcast day...''.
\end{enumerate}

The three step pipeline yields the \emph{4-Clues Labeled Image Dataset}, a collection containing 20000 images paired with structured annotations that hold bounding boxes, class names, object level captions, and a context level caption. This dataset underpins the framework because its rich annotation scheme enables the later model to learn self correction behaviour.

\subsection{Step 2: Training RT-VLM with Multi Clue Supervision}

The second phase adapts a pre trained Vision Language Model so that it can read and generate the four clue representation.

\subsubsection{Model and Architecture.}

We adopt the Llama 3.2 11B Vision Instruct model \citep{grattafiori2024llama3}. The training process relies on a parameter efficient strategy. Low Rank Adaptation (LoRA) \citep{hu2022lora} freezes the weights of the image encoder, vision adapter, and language decoder, while adding small trainable matrices to the latter two modules. In addition, nf4 quantization \citep{dettmers2023qlora} compresses the base weights so that fine tuning proceeds on a single NVIDIA 4090 GPU with 24 GB of VRAM.

\subsubsection{Instruction Tuning and Loss Function.}

Each entry of the 4-Clues dataset is transcribed into a prompt response pair. The prompt presents the image and requests the model to detect every object and to describe them following the four clue schema. The target output is a JSON document that mirrors the ground truth annotation.

\begin{table}[t]
\centering
\small
\renewcommand{\arraystretch}{1.12}
\setlength{\tabcolsep}{3pt}
\begin{tabular*}{\linewidth}{@{\extracolsep{\fill}} c  c  c  c  c @{}}
    \hline
    & \makecell{CE+LS loss} & \makecell{Schema loss} & \makecell{CB/Focal loss} & \makecell{Total loss} \\
    \hline
    Initial & 1.80 & 0.40 & 0.35 & 1.95 \\
    \hline
    Final & 0.93 & 0.14 & 0.19 & 1.00 \\
    \hline
\end{tabular*}
\caption{Decrease in training loss of the base model.}
\label{table3}
\end{table}

\begin{table}[t]
\centering
\renewcommand{\arraystretch}{1.1}
\setlength{\tabcolsep}{5pt}
\begin{tabular}{ccc}
    \hline
    & CE+LS loss & Total loss \\
    \hline
    Initial & 1.89 & 2.05 \\
    \hline
    Final & 0.98 & 1.05 \\
    \hline
\end{tabular}
\caption{Decrease in validation loss of the base model.}
\label{table4}
\end{table}

During training, the model undergoes five epochs guided by a composite loss function that is intended to balance predictive accuracy with strict conformity to the prescribed output structure (the results of which are shown in Table 3, 4). The total loss, $\mathcal L_{total}$, is therefore articulated as a weighted sum of three distinct components:

\begin{equation}
\mathcal{L}_{total} = \mathcal{L}_{CE+LS} + 0.2(\mathcal{L}_{schema} + \mathcal{L}_{class\_bal/focal})
\label{loss5}
\end{equation}

where $\mathcal L_{CE+LS}$ is the standard cross-entropy loss with label smoothing:

\begin{equation}
\mathcal{L}_{CE+LS} = -\sum_{k=1}^K y_k^{logp_k},\ p_k=\frac{e^{z_k}}{\sum_{j} e^{z_j}}
\label{loss6}
\end{equation}

$\mathcal L_{schema}$ is a schema consistency loss implicitly captured by the primary loss:

\begin{equation}
\begin{split}
\mathcal{L}_{schema} = -\sum_{t\in S}logP(t_{gold})+\lambda\frac{N_{invalid}}{|y|} \\ =-\sum_{t\in S}w_slogP(t)
\end{split}
\label{loss7}
\end{equation}

and $\mathcal L_{class\_bal/focal}$ is a class-balanced focal loss that up-weights the loss for rare or difficult-to-classify object classes:

\begin{equation}
\mathcal{L}_{class\_bal/focal} = w_y(1-p_t)^\gamma(-logp_t)
\label{loss8}
\end{equation}

This fine tuning stage converts the general purpose base VLM into the specialised RT-VLM, a model explicitly trained to inspect an image and separate it into the four evidence categories required for the subsequent stage.

\subsection{Step 3: The ``Re-Thinking'' Mechanism for Iterative Refinement}

The final component of the framework introduces a novel inference strategy called ``Re-Thinking''. The process operationalises deliberative cognition, permitting the model to transition from a rapid initial prediction toward a slower and more analytical phase of self correction.

\subsubsection{Stage 1: Evidence Generation.}

When a new zero shot image is supplied, the RT-VLM conducts an initial inference pass; it receives the image together with a ``First Prompt'' instructing it to compose a JSON object that holds the four clues, namely bounding boxes, classes, object captions, and a context caption. This first output stands as the model’s hypothesis regarding scene content and provides the raw evidence for the next stage. Nevertheless, the initial inference can incur errors because the model can be confounded by the very challenges it is intended to overcome, including covariate shift, occlusion, viewpoint variation, and class confusion.

\subsubsection{Stage 2: Guided Self Correction.}

The crucial step involves returning the data from Stage 1 into the same RT-VLM, this time accompanied by a dedicated ``Re-Thinking Prompt''. This prompt is crafted to steer the model through a process of self critique.

\begin{itemize}
\item \textbf{Re-Thinking Prompt:} Your first inference results can be erroneous because of (1) covariate shifts, (2) occlusion, (3) orientation or viewpoint variation, or (4) class confusion for some objects in the input image. If, after reflection, you consider that your first inference results (boxes, classes, object$\_$captions, context$\_$caption) for every object in the image lack accuracy for any of the four issues listed, or that certain objects were not inferred at all because of those issues and you therefore hold low confidence in your primary inference, review and correct or add to the ``boxes'', ``classes'', and ``object$\_$captions'' fields for such objects and, if needed, adjust the ``context$\_$caption'' field as well. The criteria for review, additions, and corrections are [1] the information in the context$\_$caption field from the first inference together with the information in the [2] boxes field, [3] classes field, and [4] object$\_$captions field of other correctly inferred objects that do not require revision. In other words, use the global contextual information at image level, namely which objects were incorrectly inferred or entirely missed because of (1) covariate shifts, (2) occlusion, (3) viewpoint changes, or (4) class confusion, and the object level context of correctly inferred neighbours as clues to revisit the problematic or missing objects and to produce the correct inference once again. Output your secondary inference with final corrections and any necessary additions as shown above.
\end{itemize}

This prompt invites the model to serve as its own reviewer, leveraging previously generated evidence to check consistency. The successful execution of the ``Re-Thinking'' stage results directly from the earlier steps; the model carries out structured reasoning precisely because it was fine tuned on the 4-Clues dataset. This tight link among dataset design, model training, and inference mechanism underpins the conceptual strength of the RT-VLM framework. The output from Stage 2 is a refined, more accurate, and more robust set of object recognitions.

\section{Experimental Evaluation}

\subsection{Experimental Setup}

To rigorously evaluate the effectiveness of the RT-VLM framework and the ``Re-Thinking'' strategy, we performed experiments across diverse benchmarks; the study not only reports overall scores, it also isolates the influence of each component.

\subsubsection{Datasets.}

A wide range of datasets was used to test the models under different conditions:

\begin{itemize}[leftmargin=1em,labelsep=0.5em]
    \item \textbf{General Performance:} A custom-generated test dataset (5,000 synthetic images designed to be challenging) and the standard MS-COCO val2017 dataset \citep{lin2014microsoft} were used for object recognition. \\ Flickr30k Entities \citep{plummer2015flickr30k} and MS-COCO Captions \citep{chen2015microsoft} were used for object-level and context-level captioning, respectively.
    \item \textbf{Robustness Benchmarks:} To specifically test resilience to domain shift, four sets of specialized benchmarks were employed:
        \begin{itemize}
            \item \textbf{Covariate Shift:} COCO-C and Pascal-C \citep{hendrycks2019benchmarking}, which apply 15 types of synthetic corruptions (e.g., blur, noise, weather effects) to the COCO and PASCAL-VOC \citep{everingham2010pascal} datasets.
            \item \textbf{Orientation and Viewpoint Variation:} \\ Pascal3D+ \citep{xiang2014beyond} and ObjectNet3D \citep{xiang2016objectnet3d}, which feature objects in a wide range of challenging 3D poses and viewpoints.
            \item \textbf{Occlusion:} Occluded PASCAL3D+, a dataset where objects are synthetically occluded at varying levels of severity \citep{xiang2014beyond}.
            \item \textbf{Class Confusion:} ImageNet-A \\ \citep{hendrycks2021natural}, a collection of real-world images that are consistently misclassified by standard models, serving as a test for handling natural adversarial examples.
        \end{itemize}
\end{itemize}

\subsubsection{Metrics and Baselines.}

Task specific metrics were adopted in line with prior work. Object recognition used mean Average Precision, Precision, Recall, F1 Score, and mean Intersection over Union as defined in Equations 9 to 11 \citep{everingham2010pascal,vanrijsbergen1979information}. Captioning quality was quantified with CIDEr \citep{vedantam2015cider} and BLEU-4 \citep{papineni2002bleu}. Classification accuracy was reported as Top 1 and Top 5. To reveal the value added by each design choice, we compared four model variants within an ablation study:

\begin{equation}
mAP=\frac{1}{N}\sum_{i=1}^NAP_i
\label{metric1}
\end{equation}

\begin{equation}
Precision=\frac{TP}{TP+FP},\ Recall=\frac{TP}{TP+FN}
\label{metric2}
\end{equation}

\begin{equation}
F1=2\cdot\frac{Precision\cdot Recall}{Precision+Recall}
\label{metric3}
\end{equation}

\begin{enumerate}
    \item \textbf{Base VLM (A):} The pre-trained \\ Llama-3.2-11B-Vision-Instruct model \citep{grattafiori2024llama3} without any fine-tuning.
    \item \textbf{Base VLM + Re-Thinking (B):} The base model using the two-stage Re-Thinking inference.
    \item \textbf{RT-VLM (C):} The fine-tuned model using only a single-pass inference.
    \item \textbf{RT-VLM + Re-Thinking (D):} The full proposed system, combining the fine-tuned model with the two-stage Re-Thinking inference.
\end{enumerate}

\subsection{Analysis of Core Performance and Robustness}

Experimental evidence consistently indicates that the full RT-VLM framework secures top scores across the entire evaluation suite. The collected observations further reveal subtle dependencies linking domain oriented training with the reasoning module. A concise digest of the most salient outcomes is presented in the subsequent tables (refer to Tables~5 through~12 of the source material for full details).

\setlength{\tabcolsep}{2.8pt}
\renewcommand{\arraystretch}{1.1}

\newcolumntype{M}{>{\centering\arraybackslash}p{66pt}}
\newcolumntype{C}{>{\centering\arraybackslash}X}

\begin{table}[!htbp]
  \centering\small
  \caption{Performance evaluation of model variants on object\\-recognition tasks using custom-generated test dataset (A: Base VLM, B: Base VLM + Re-Thinking, C: RT-VLM, D: RT-VLM + Re-Thinking).}
  \label{table5}
  \begin{tabularx}{\linewidth}{@{}M C C C C@{}}
    \toprule
    Metric & A & B & C & D \\ \midrule
    mAP@0.5        & 0.34 & 0.34 & 0.68 & \textbf{0.69} \\
    mAP@0.5:0.95   & 0.16 & 0.15 & \textbf{0.44} & \textbf{0.44} \\
    IoU Mean       & 0.31 & 0.33 & \textbf{0.52} & \textbf{0.52} \\
    Precision      & 0.31 & 0.34 & 0.68 & \textbf{0.70} \\
    Recall         & 0.27 & 0.29 & 0.65 & \textbf{0.66} \\
    F1 Score       & 0.30 & 0.30 & 0.66 & \textbf{0.70} \\ \bottomrule
  \end{tabularx}

\end{table}

\setlength{\tabcolsep}{2.8pt}
\renewcommand{\arraystretch}{1.1}

\begin{table}[!htbp]
  \centering\small
  \caption{Performance evaluation of model variants on object\\-recognition tasks using COCO val2017 dataset (A: Base VLM, B: Base VLM + Re-Thinking, C: RT-VLM, D: RT-VLM + Re-Thinking).}
  \label{table6}
  \begin{tabularx}{\linewidth}{@{}M C C C C@{}}
    \toprule
    Metric & A & B & C & D \\ \midrule
    mAP@0.5        & 0.65 & 0.66 & \textbf{0.76} & \textbf{0.76} \\
    mAP@0.5:0.95   & 0.42 & 0.42 & 0.51 & \textbf{0.52} \\
    IoU Mean       & 0.56 & 0.56 & 0.63 & \textbf{0.65} \\
    Precision      & 0.71 & 0.73 & 0.80 & \textbf{0.84} \\
    Recall         & 0.63 & 0.63 & \textbf{0.72} & \textbf{0.72} \\
    F1 Score       & 0.67 & 0.68 & 0.74 & \textbf{0.77} \\ \bottomrule
  \end{tabularx}  
\end{table}

\begin{table}[!htbp]
  \centering\small
   \caption{Performance evaluation of model variants on object\\-level captioning tasks using test dataset and Flickr30k Entities dataset (A: Base VLM, B: Base VLM + Re-Thinking, C: RT-VLM, D: RT-VLM + Re-Thinking).}
  \label{table7}
  \begin{tabularx}{\linewidth}{@{}M M C C C C@{}}
    \toprule
    Dataset & Metric & A & B & C & D \\ \midrule
    Test Dataset & CIDEr & 68.9 & 68.2 & 149.5 & \textbf{151.7} \\
    Test Dataset & BLEU-4 & 20.2 & 20.0 & 42.8 & \textbf{45.1} \\
    Flickr30k Entities & CIDEr & 129.7 & 129.7 & 136.9 & \textbf{137.4} \\
    Flickr30k Entities & BLEU-4 & 35.7 & 35.9 & 38.9 & \textbf{39.4} \\ \bottomrule
  \end{tabularx} 
\end{table}

\begin{table}[!htbp]
  \centering\small
  \caption{Performance evaluation of model variants on contextual (full-image) captioning tasks using test dataset and MS-COCO Captions dataset (A: Base VLM, B: Base VLM + Re-Thinking, C: RT-VLM, D: RT-VLM + Re-Thinking).}
  \label{table8}
  \begin{tabularx}{\linewidth}{@{}M M C C C C@{}}
    \toprule
    Dataset & Metric & A & B & C & D \\ \midrule
    Test Dataset & CIDEr & \textbf{131.9} & 131.6 & 128.4 & 128.4 \\
    Test Dataset & BLEU-4 & \textbf{38.8} & \textbf{38.8} & 37.1 & 37.2 \\
    MS-COCO Captions & CIDEr & 137.7 & \textbf{137.8} & 131.6 & 131.9 \\
    MS-COCO Captions & BLEU-4 & \textbf{39.1} & \textbf{39.1} & 37.4 & 36.8 \\ \bottomrule
  \end{tabularx}  
\end{table}

\begin{table}[!htbp]
  \centering\small
  \caption{Performance evaluation of model variants on covariate shift using COCO-C dataset and Pascal-C dataset (A: Base VLM, B: Base VLM + Re-Thinking, C: RT-VLM, D: RT-VLM + Re-Thinking).}
  \label{table9}
  \begin{tabularx}{\linewidth}{@{}M M C C C C@{}}
    \toprule
    Dataset & Metric & A & B & C & D \\ \midrule
    COCO-C & mAP@0.5 & 0.48 & 0.50 & 0.70 & \textbf{0.75} \\
    COCO-C & mAP@0.5:0.95 & 0.29 & 0.34 & 0.46 & \textbf{0.50} \\
    Pascal-C & mAP@0.5 & 0.53 & 0.57 & 0.72 & \textbf{0.75} \\
    Pascal-C & mAP@0.5:0.95 & 0.34 & 0.38 & 0.47 & \textbf{0.51} \\ \bottomrule
  \end{tabularx}  
\end{table}

\begin{table}[!htbp]
  \centering\small
  \caption{Performance evaluation of model variants on orientation/viewpoint variation using Pascal3D+ dataset and ObjectNet3D dataset (A: Base VLM, B: Base VLM + Re-Thinking, C: RT-VLM, D: RT-VLM + Re-Thinking).}
  \label{table10}
  \begin{tabularx}{\linewidth}{@{}M M C C C C@{}}
    \toprule
    Dataset & Metric & A & B & C & D \\ \midrule
    Pascal3D+ & mAP@0.5 & 0.57 & 0.61 & 0.74 & \textbf{0.80} \\
    Pascal3D+ & mAP@0.5:0.95 & 0.35 & 0.36 & 0.48 & \textbf{0.52} \\
    ObjectNet3D & mAP@0.5 & 0.51 & 0.53 & 0.70 & \textbf{0.75} \\
    ObjectNet3D & mAP@0.5:0.95 & 0.26 & 0.26 & 0.45 & \textbf{0.47} \\ \bottomrule
  \end{tabularx}  
\end{table}

\begin{table}[!htbp]
  \centering\small
  \caption{Performance evaluation of model variants on occlusion using OccludedPASCAL3D+ dataset (A: Base VLM, B: Base VLM + Re-Thinking, C: RT-VLM, D: RT-VLM + Re-Thinking).}
  \label{table11}
  \begin{tabularx}{\linewidth}{@{}M M C C C C@{}}
    \toprule
    Dataset & Metric & A & B & C & D \\ \midrule
    Occluded PASCAL3D+ & mAP@0.5 & 0.15 & 0.21 & 0.27 & \textbf{0.34} \\
    Occluded PASCAL3D+ & mAP@0.5:0.95 & 0.09 & 0.12 & 0.22 & \textbf{0.24} \\ \bottomrule
  \end{tabularx}  
\end{table}

\begin{table}[!htbp]
  \centering\small
  \caption{Performance evaluation of model variants on class confusion using ImageNet-A dataset (A: Base VLM, B: Base VLM + Re-Thinking, C: RT-VLM, D: RT-VLM + Re-Thinking).}
  \label{table12}
  \begin{tabularx}{\linewidth}{@{}M M C C C C@{}}
    \toprule
    Dataset & Metric & A & B & C & D \\ \midrule
    ImageNet-A & Top-1 Accuracy (\%) & 64.81 & 64.81 & 72.74 & \textbf{74.11} \\
    ImageNet-A & Top-5 Accuracy (\%) & 84.55 & 84.57 & 88.14 & \textbf{89.87} \\ \bottomrule
  \end{tabularx}  
\end{table}

\subsubsection{Key Finding 1: The Synergy Between Specialized Training and Re-Thinking is Critical.}

The analysis shows that the Re-Thinking mechanism, although only marginally beneficial in isolation, achieves a marked gain when paired with the specialized RT-VLM.  
The gain becomes clear once the prompt is examined, since it directs the model to infer relationships among bounding boxes, object attributes, and the wider scene context.  
The Base VLM is a competent generalist; however, it lacks explicit exposure to the 4-Clues representation, therefore its capacity to generate or parse that structure is limited.  
The RT-VLM, in contrast, was fine tuned on the 4-Clues dataset and thus assimilated that representation.  
The interplay between task specific training and the deliberative procedure consequently yields a stronger result.

\subsubsection{Key Finding 2: Specialization for Object Centric Tasks Involves a Trade off.}

Although RT-VLM produces strong results on object centric evaluations, empirical data disclose a distinct trade off.  
In contextual captioning on MS~COCO, the fine tuned RT-VLM performs below the original Base VLM.  
Such an outcome is anticipated and aligns with the theory of specialization \citep{houlsby2019parameter}.  
The fine tuning routine guided the parameter space towards recognising and enumerating discrete objects that fit the 4-Clues specification.  
That targeted emphasis seems to have reduced the capacity to construct coherent scene wide captions with fluent narrative flow.

\subsubsection{Key Finding 3: Superior Robustness Across All Four Targeted Challenges.}

The most prominent observation concerns the framework itself, which records consistent and substantial gains in robustness across the four examined domain shift conditions.  

Across every scenario investigated, the joint application of 4-Clues training and Re-Thinking inference acts as an effective countermeasure against the limitations observed in baseline models, thereby supporting the central hypothesis of this study.

\section{Conclusion and Future Work}

In this study, we presented RT-VLM, a framework that addresses domain shift in object recognition by fine tuning a VLM on a novel synthetic dataset enriched with ``4-Clues'' annotations and by applying a two stage Re-Thinking inference step for self correction. Extensive experiments confirmed our central claim, demonstrating that richer supervision combined with a deliberate reasoning procedure delivers stronger resilience to covariate shift, orientation or viewpoint variation, occlusion, and class confusion. These findings indicate an encouraging path toward dependable reasoning oriented AI. Although the approach relies on synthetic data, forthcoming work can broaden the framework by exploring multi-hop reasoning, integrating external knowledge sources, and constructing a fully end to end architecture, thereby advancing the frontier of robust AI.

\bigskip

\bibliography{aaai2026}

\end{document}